\documentclass{article} 
\usepackage{iclr2023_conference_tinypaper,times}

\usepackage{amsmath,amsfonts,bm}

\def\eqref#1{equation~\ref{#1}}

\def\1{\bm{1}}

\DeclareMathAlphabet{\mathsfit}{\encodingdefault}{\sfdefault}{m}{sl}
\SetMathAlphabet{\mathsfit}{bold}{\encodingdefault}{\sfdefault}{bx}{n}

\usepackage{float}
\usepackage{hyperref}
\usepackage{url}
\iclrfinalcopy
\usepackage{graphicx}
\graphicspath{ {./images/} }
\usepackage{amssymb}

\usepackage[utf8]{inputenc} 
\usepackage[T1]{fontenc}    
\usepackage{hyperref}       
\usepackage{url}            
\usepackage{booktabs}       
\usepackage{amsfonts}       
\usepackage{nicefrac}       
\usepackage{microtype}      
\usepackage{lipsum}
\usepackage{fancyhdr}       
\usepackage{graphicx}       
\graphicspath{{media/}}     
\usepackage{tablefootnote}
\usepackage{subcaption}
\usepackage{float}
\usepackage{hyperref}
\usepackage{amsmath}
\makeatletter
\newcommand{\printfnsymbol}[1]{%
  \textsuperscript{\@fnsymbol{#1}}%
}
\makeatother
\DeclareUnicodeCharacter{2061}{}

\title{An Empirical Study of the Effect of Background Data Size on the Stability of SHapley Additive exPlanations (SHAP) for Deep Learning Models}

\author{Han Yuan$^*$$^,$$^\infty$ \& Mingxuan Liu$^*$ \& Lican Kang \\
Duke-NUS Medical School \\
National University of Singapore \\
\And
\And
Chenkui Miao \\
The First School of Clinical Medicine \\
Nanjing Medical University \\
\And
Ying Wu \\
School of Statistics and Data Science \\
Nankai University \\
}

\begin{document}
\maketitle

\def\thefootnote{*}\footnotetext{These authors contributed equally; $^\infty$yuan.han@u.duke.nus.edu}
\def\thefootnote{\arabic{footnote}}
\begin{abstract}
Nowadays, the interpretation of why a machine learning (ML) model makes certain inferences is as crucial as accuracy. Some ML models like the decision tree possess inherent interpretability that can be directly comprehended by humans. Others like deep learning (DL) models, however, rely on external methods to uncover the deduction mechanism. SHapley Additive exPlanations (SHAP) is one such external method, which requires a background dataset when interpreting DL models. Generally, a background dataset consists of instances randomly sampled from the training dataset. However, the sampling size and its effect on SHAP remain to be unexplored. Our empirical study on the MIMIC-III dataset shows that the two core explanations - SHAP values and variable rankings fluctuate when using different background datasets acquired from random sampling, indicating that users cannot unquestioningly trust the interpretation from SHAP. Luckily, such fluctuation decreases with the increase in the background dataset size. Also, we notice a U-shape in the stability assessment of SHAP variable rankings, demonstrating that SHAP is more reliable in ranking the most and least important variables compared to moderately important ones. Overall, our results suggest that users should consider how background data affects SHAP results, with improved SHAP stability as the background sample size increases. Code is publicly accessible\footnote{\url{https://github.com/Han-Yuan-Med/shap-bg-size}}.
\end{abstract}

\section{Introduction}
Nowadays, interpreting model inference is becoming as important as techniques for enhancing model accuracy \citep{lauritsen2020explainable}. Some machine learning (ML) models like the decision tree can be easily understood by humans while others like ANNs are too complex to be interpreted even by experts \citep{NIPS2017_8a20a862,XIE2022103980}. To address this problem, researchers proposed various explanations: instance-level and model-level. The instance-level explanation typically provides feature contribution analyses for a single prediction, and the model-level explanation provides feature importance across all predictions.

SHAP provides both instance and model-level explanations through SHAP values and variable rankings \citep{NIPS2017_8a20a862}. SHAP values are the direct production from SHAP calculations while variable rankings are measured by the sum of each variable’s absolute SHAP values across all instances \citep{lundberg2020local}. To eliminate computing complexity, the SHAP package\footnote{\url{https://pypi.org/project/shap/}} includes various explainers for different ML models. DeepExplainer is an efficient explainer for deep learning (DL) models and requires a background dataset to serve as a prior expectation for the instances to be explained.

While the official SHAP documentation\footnote{\url{https://shap.readthedocs.io/en/latest/}} suggests 100 randomly drawn samples from the training data as an adequate background dataset, other studies employed different sampling sizes \citep{van2020volumetric,giese2021retention,kaufmann2021efficient}. This conflict raises questions: What is the recommended background data size on SHAP explanations and is there an impact of using different data sizes? In a pilot experiment, we found that both instance-level SHAP values and model-level variable rankings fluctuate when we adopt random sampling to obtain background datasets with small sizes. To quantify such fluctuation and better answer the questions above, we conducted an empirical study with different background dataset sizes applied to interpret three-layer ANNs (See Appendix \ref{appnedixa}). We then quantified the effect on instance-level explanations (SHAP values). Lastly, we used an exact (BLEU score \citep{papineni2002bleu}) and fuzzy (Jaccard index \citep{Fletcher_Islam_2018}) approach for evaluating model-level explanations (variable rankings).

\section{Method}
\subsection{SHAP values and variable rankings}
SHAP provides instance-level and model-level explanations by SHAP value and variable ranking. In a binary classification task (the label is 0 or 1), the inputs of an ANN model are variables $ var_{i,j} $ from an instance $ D_i $, and the output is the prediction probability $ P_i $ of $ D_i $ of being classified as label 1. In general, we are interested in interpreting a stack of instances $D$ at both the instance and model levels.

\begin{equation}
D=\left\{D_{i}\right\}, i=1, \ldots, N
\label{formula 1}
\end{equation}

\begin{equation}
D_i=\left\{var_{i,j}\right\}, j=1, \ldots, V
\label{formula 2}
\end{equation}

\begin{equation}
P_i=ANN(D_i), i=1, \ldots, N
\label{formula 3}
\end{equation}

where $ i $ represents the $i$-th instance in $D$ and $j$ stands for the $j$-th variable in $D_i$.

DeepExplainer, the function for SHAP calculations in an ANN, provides instance-level explanations of $P_i$ through the contribution ($shap_{i,j,bg}$) of each variable ($var_{i,j}$) to the prediction deviation from a prior $P_{bg}$, which is the probability expectation of samples in the background dataset $bg$.
\begin{equation}
P_{i}-P_{bg}=\sum_{j=1}^{V} shap_{i, j, b g}
\label{formula 4}
\end{equation}
With the instance-level $shap_{i,j,bg}$,  we then compute the model-level variable importance $I_{j,bg}$ \citep{NIPS2017_8a20a862} as follows:
\begin{equation}
    I_{j, b g}=\sum_{i=1}^{N}\left|shap_{i, j, b g}\right|
    \label{formula 5}
\end{equation}
where absolute SHAP values for a particular variable are summed over all instances in $D$. Based on variable importance, users obtain the variable rankings effortlessly (larger $I_{j,bg}$, higher importance and ranking).

As shown in formula~\ref{formula 4} and \ref{formula 5}, a background dataset is a prerequisite for DeepExplainer and its application on ANNs interpretation. Generally, a background dataset consists of instances randomly sampled from the training data \citep{NIPS2017_8a20a862}. However, the use of small background sample sizes potentially causes fluctuation of SHAP explanations, which may affect  users’ trust in the one-shot SHAP explanations.

While the statistical variance is well suited for evaluating the fluctuations of SHAP values, we need more refined measures for assessing changes in variable rankings. Specifically, as changes in variable rankings can be seen as equivalent to word order changes, we adopt the BLEU assessment, typically used to compare the word order of a translated text with regard to a reference translation, for measuring differences in variable rankings. Another potential evaluation comes from similarity computation in set theory: the Jaccard index.

\subsection{Fluctuation quantification of variable ranking}
\subsubsection{BLEU for exact fluctuation evaluations}

BLEU \citep{papineni2002bleu} aims for evaluating machine translation by comparing n-gram matches between the candidate translation and the reference translation. An n-gram is a contiguous sequence of n items from a given text. The first step in BLEU is to compute the n-gram matches precision score $p_n$: the matched n-gram counts in the candidate translation over the number of candidate n-grams in the reference translation. The second step is to multiply $p_n^{\frac{1}{2^n}} (1 \leq n \leq k $, $n$ is an integer$)$ with penalty terms: $\exp(\min⁡(0,1-\frac{l_c}{l_r}))$, where $k$ is the maximum number of words in the matched subsequence, and $l_c$ and $l_r$ stand for lengths of the candidate translation and the reference translation, respectively.
\begin{equation}
\mathrm{BLEU}=\exp \left(\min \left(0,1-\frac{l_{c}}{l_{r}}\right)\right) \prod_{n=1}^{k} p_n^{\frac{1}{2^{n}}}
\label{formula 6}
\end{equation}
As stated above, if we treat one variable ranking as “the candidate translation” and the other as “the reference translation”, we can also compute the difference between these two rankings. The formula~\ref{formula 6} is modified to suit our application: (1) Since the ranking sequences are of equal length and contain the same elements, the $\exp(\min⁡(0,1-\frac{l_c}{l_r}))$ and $p_1$ are always 1; (2) Only two-grams BLEU is used to quantify the relevant sequence between two rankings; (3) Considering that two-grams BLEU might fail to detect the drastic order change of two-grams units, we proposed quartile-based computation where the variable ranking is split into four quartiles, $p_{2,q}^{\frac{1}{4}}$ is calculated for each quartile $q\, (q=1,2,3,4)$, and the average value $\mathrm{BLEU_Q}$ serves as the final assessment value.
\begin{equation}
\mathrm{BLEU}_{\mathrm{Q}}=\frac{1}{4} \times \sum_{q=1}^{4} p_{2, q}^{\frac{1}{4}}
\label{formula 7}
\end{equation}
We use a fictive example to clarify $\mathrm{BLEU_Q}$ (See Table~\ref{tab:table1}). There are two rankings: Ranking \#1 [a, b, c, d, e, f, g, h, i, j, k, l] and Ranking \#2 [a, c, b, d, e, f, h, g, i, j, k, l]. We first split the two rankings into four parts. For example, Ranking \#1 is split into [a, b, c], [d, e, f], [g, h, i], and [j, k, l]. Then we calculate two-grams in each quartile of these two rankings. Ranking \#1’s first part is [a, b, c] and the corresponding two-grams are [a, b] and [b, c]. With two-grams in each quartile, we can easily obtain the matched two-grams counts and the number of two-grams. Finally, the $\mathrm{BLEU_Q}$ here is $\frac{1}{4} \times (0^\frac{1}{4}+1^\frac{1}{4}+0^\frac{1}{4}+1^\frac{1}{4})=0.25$ according to formula~\ref{formula 7}.

\begin{table}[h]
\caption{Two-grams and the corresponding precision score of Ranking \#1 and \#2}
\centering
    \begin{tabular}{lllccc}
    \hline
    Quartile & Ranking \#1          & Ranking \#2           & Matched Num & Reference Num & Precision Score \\
    \hline
    0-25\%   & {[}a, b{]}, {[}b, c{]} & {[}a, c{]}, {[}c, b{]} & 0           & 2             & 0/2 = 0 \\
    25-50\%  & {[}d, e{]}, {[}e, f{]} & {[}d, e{]}, {[}e, f{]}  & 2           & 2             & 2/2 = 1 \\
    50-75\%  & {[}g, h{]}, {[}h, i{]} & {[}h, g{]}, {[}g, i{]}  & 0           & 2             & 0/2 = 0 \\
    75-100\% & {[}j, k{]}, {[}k, l{]} & {[}j, k{]}, {[}k, l{]}  & 2           & 2             & 2/2 = 1 \\
    \hline
    \end{tabular}
\label{tab:table1}
\end{table}

\subsubsection{Jaccard index for fuzzy fluctuation evaluations}
In contrast to $\mathrm{BLEU_Q}$ which focuses on the exact match of two-gram units, Jaccard index \citep{Fletcher_Islam_2018}, defined as the size of the intersection divided by the size of the union of the sample sets, is used for evaluating the fuzzy similarity between sample sets. Given that any ranking in our study contains the same variables, the Jaccard index cannot be directly used.

Like $\mathrm{BLEU_Q}$, we propose the quartile-based $\mathrm{Jaccard_Q}$, wherein the variable ranking is split into four quartiles, $Jaccard_q$ is calculated in each quartile $q\, (q=1,2,3,4)$, and the average value across all quartiles works as a final assessment of the fluctuation of variable rankings.
\begin{equation}
\mathrm{Jaccard}_{\mathrm{Q}}=\frac{1}{4} \times \sum_{q=1}^{4} {Jaccard}{ }_{q}
\label{formula 8}
\end{equation}
The same fictive sample is used for clarification (See Table~\ref{tab:table2}): We first split the two rankings into four parts, then compute the intersection, union number, and Jaccard index in each quartile, and finally obtain the mean value of all Jaccard indexes $\frac{1}{4} \times (1+1+1+1)=1$.

\begin{table}[h]
\caption{Variable subsets and corresponding Jaccard index of Ranking \#1 and \#2}
\centering
\begin{tabular}{lccccc}
\toprule
Quartile & Ranking \#1 & Ranking \#2 & Intersection Num & Union Num & Jaccard index \\
\hline
0-25\%   & a, b, c       & a, c, b       & 3                & 3         & 3/3 = 1       \\
25-50\%  & d, e, f       & d, e, f       & 3                & 3         & 3/3 = 1       \\
50-75\%  & g, h, i       & h, g, i       & 3                & 3         & 3/3 = 1       \\
75-100\% & j, k, l       & j, k, l       & 3                & 3         & 3/3 = 1       \\
\hline
\end{tabular}
\label{tab:table2}
\end{table}

\subsection{Dataset and model architecture}
We implemented an empirical study of SHAP stability using a de-identified intensive care unit dataset. This dataset includes 44,918 admission episodes (including 3,958 positive episodes, defined as admissions within patient mortality) of the Beth Israel Deaconess Medical Center \citep{johnson2016mimic}. We randomly separated the data set into development and explanation sets. The development set consisted of 31,442 (70\%) patients, and the explanation set was made up of 13,476 (30\%) patients. The development set was used to develop the ANN and to generate background datasets. The explanation set was put aside to be interpreted by SHAP. The variables to be ranked included heart rate, age, respiration rate, systolic blood pressure, diastolic blood pressure, mean arterial pressure, white blood cell count, platelet count, glucose, sodium, lactate, bicarbonate, blood urea nitrogen, creatinine, chloride. An ANN with three layers was used as a backbone model in this study because no substantial gain was observed with more layers. The ANN was made up of 2 hidden layers with 128 and 64 rectified linear units respectively and 1 output layer using sigmoid activation. 

\subsection{Empirical study setting}
We varied background data size from 100 to 1,000 and performed 100 simulations under each background data size. In each simulation, a background dataset with a fixed size was sampled from the training dataset. Then SHAP values and variable rankings are calculated on the explanation set. After 100 simulations, we obtained 100 SHAP values for each variable in a single instance and applied statistical variance to depict the fluctuation of SHAP values in this instance: For variable $var_j$, its variance sum is $\sum_{i=1}^{N} \frac{1}{99} \sum_{bg=1}^{100}\left(\left|shap_{i,j,bg}\right|-\frac{1}{100} \sum_{bg=1}^{100}\left|shap_{i,j,bg}\right|\right)^{2}$. Also, we received $p=100$ variable rankings and $C_p^2$ different pairs of rankings in the model level. $\mathrm{BLEU_Q}^k$  and $\mathrm{Jaccard_Q}^k$ represents the quartile-based $\mathrm{BLEU_Q}$ and $\mathrm{Jaccard_Q}$ index of the $k$-th pair, respectively. Then the mean of $\mathrm{BLEU_Q}^k$ and $\mathrm{Jaccard_Q}^k$ across all pairs were calculated to assess the fluctuation of variable rankings. All computations were carried out using PyTorch version 1.6.0, Python version 3.8, and R version 4.0.3.
\begin{equation}
\mathrm{Mean\;BLEU}=\sum_{k=1}^{C_{p}^{2}}\left(\mathrm{BLEU}_{\mathrm{Q}}^{k}\right); \mathrm{Mean\;Jaccard}=\sum_{k=1}^{C_{p}^{2}}\left(\mathrm{Jaccard}_{\mathrm{Q}}^{k}\right)
\label{formula 9}
\end{equation}

\section{Experiments}
Using MIMIC-III data, we evaluated the impact of the background dataset size on both instance-level and model-level SHAP explanations. The fluctuation of instance-level SHAP values was assessed using statistical variance, while the instability of model-level variable rankings was evaluated using the proposed $\mathrm{Mean\;BLEU}$ and $\mathrm{Mean\;Jaccard}$ measures.

\subsection{Fluctuation of SHAP values}
The fluctuation of instance-level SHAP explanations originates from changes of SHAP values, as assessed by a statistical variance measure. We observe that the variance sum per variable across instances in the explanation set  decreases as the background sample size increases (Table~\ref{tab:table3}).

\begin{table}[h]
\caption{The mean variance of SHAP values across all observations}
\centering
\begin{tabular}{lccc}
\toprule
Variables                & \begin{tabular}[c]{@{}l@{}}Variance sum\\ (sample size=100)\end{tabular} & \begin{tabular}[c]{@{}l@{}}Variance sum\\ (sample size=500)\end{tabular} & \begin{tabular}[c]{@{}l@{}}Variance sum\\ (sample size=1000)\end{tabular} \\
\midrule
Age                      & 153.31                                                                                                                  & 22.16                                                          & 11.21                                                           \\
Heart rate               & 26.71                                                                                                                    & 8.85                                                           & 5.20                                                             \\
Systolic blood pressure  & 271.31                                                                                                                   & 44.27                                                          & 24.13                                                           \\
Diastolic blood pressure & 74.64                                                                                                                    & 11.22                                                          & 5.52                                                            \\
Arterial pressure        & 169.92                                                                                                                   & 22.91                                                          & 13.32                                                           \\
Respiration rate         & 33.08                                                                                                                     & 5.75                                                           & 2.85                                                            \\
Temperature              & 0.50                                                                                                                      & 0.10                                                           & 0.05                                                            \\
SpO2                     & 1.70                                                                                                                      & 0.21                                                           & 0.13                                                            \\
Glucose                  & 72.18                                                                                                                    & 16.35                                                          & 7.32                                                            \\
Anion gap                & 5.64                                                                                                                     & 1.54                                                           & 0.77                                                            \\
Bicarbonate              & 10.17                                                                                                                     & 2.02                                                           & 1.11                                                            \\
Creatinine               & 56.95                                                                                                                    & 7.92                                                           & 3.82                                                            \\
Chloride                 & 67.52                                                                                                                    & 7.85                                                           & 3.17                                                            \\
Lactate                  & 22.63                                                                                                                    & 6.93                                                           & 3.05                                                            \\
Hemoglobin               & 1.56                                                                                                                      & 0.18                                                           & 0.10                                                             \\
Hematocrit               & 3.58                                                                                                                      & 0.35                                                           & 0.18                                                            \\
Platelet                 & 325.33                                                                                                                  & 49.46                                                          & 23.21                                                           \\
Potassium                & 3.71                                                                                                                      & 0.86                                                           & 0.47                                                            \\
Blood urea nitrogen      & 212.63                                                                                                                   & 30.46                                                          & 15.60                                                            \\
Sodium                   & 60.20                                                                                                                     & 6.54                                                           & 2.65                                                            \\
White blood cells        & 83.88                                                                                                                    & 10.40                                                          & 5.71                                                            \\
\bottomrule
\end{tabular}
\label{tab:table3}
\end{table}

\subsection{Fluctuation of variable rankings}
\begin{figure}[h]
    \centering
    \begin{subfigure}[b]{1\textwidth}
           \centering
           \includegraphics[width=\textwidth]{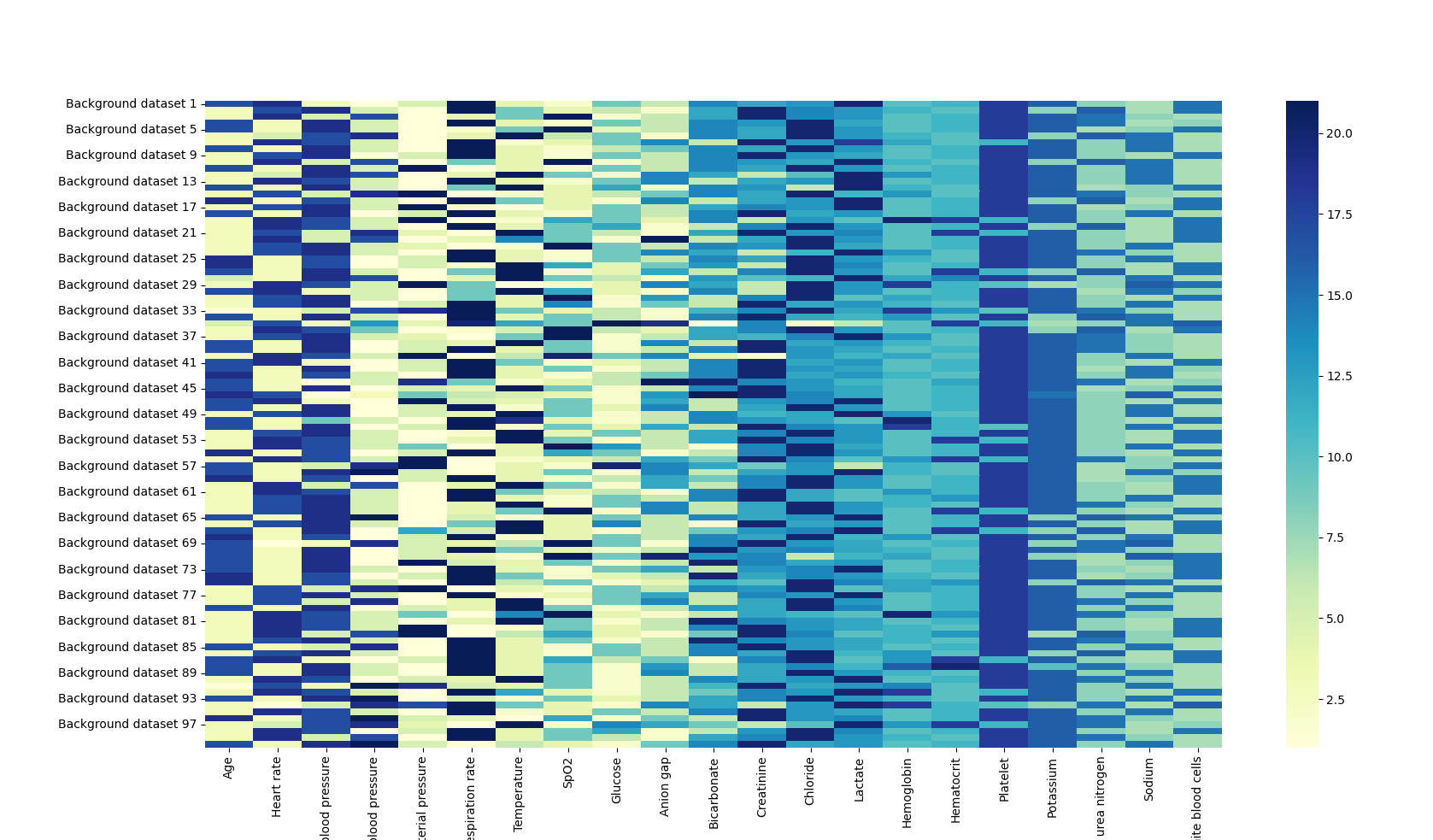}
            \caption{The variable rankings using 100 background dataset samples.}
            \label{fig:a}
    \end{subfigure}
    \begin{subfigure}[b]{1\textwidth}
            \centering
            \includegraphics[width=\textwidth]{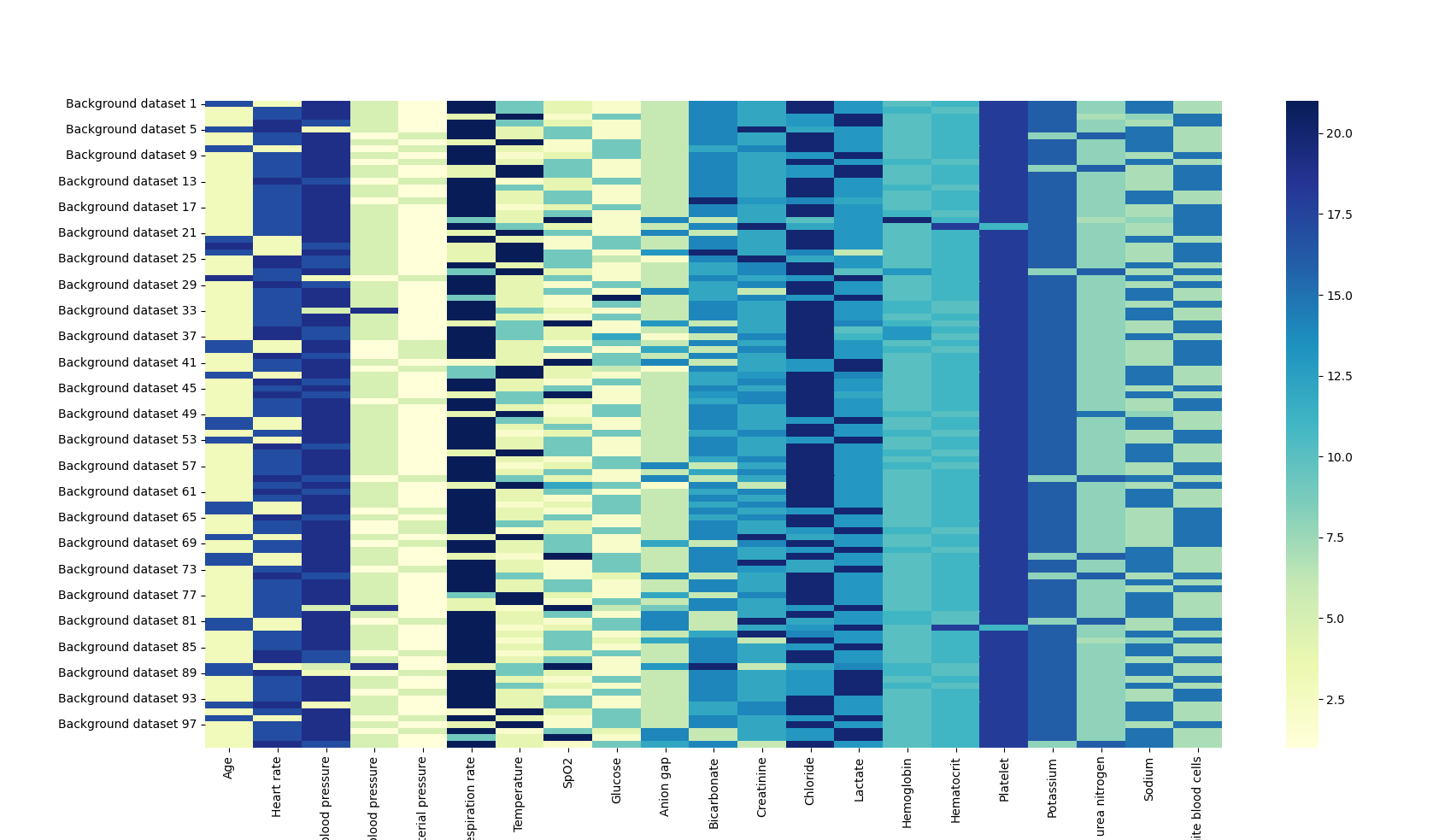}
            \caption{The variable rankings using 1,000 background dataset samples.}
            \label{fig:b}
    \end{subfigure}
    \caption{Fluctuation of variable rankings using background sizes of 100 and 1,000. Based on 100 simulations, we obtained 100 variable rankings for each of the two background sizes. Each row corresponds to one simulation; each column represents a variable’s ranking order across the 100 simulations. The color bar on the right indicates the variable ranking sequence (blue means low rankings and yellow stands for high rankings). The small color blocks in each column of sub-figure (b) change less than these of (a), indicating smaller volatilities.}
\label{fig:fig1}
\end{figure}

Fluctuation of instance-level SHAP values may also indicate unstable model-level variable rankings. Figure~\ref{fig:fig1} visualizes the simulation results with background dataset sizes of 100 and 1,000, respectively. We observe more stable rankings with the larger dataset.

Though better stability was observed in larger background data, quantitative assessment is still necessary for accurate analyses. We utilized the quartile-based BLEU and Jaccard index to quantify the exact and fuzzy stabilities. Tables~\ref{tab:table4} and \ref{tab:table5} demonstrate that the pairwise analyses of the different rankings resulted in improved BLEU and Jaccard scores when the background dataset size increased. While the Jaccard index was close to 0.9 across all four quartiles, the BLEU results indicate that the relevant variable orders fluctuate even using many background samples, with a score of 0.644 for 1000 background samples. Interestingly, the BLEU and Jaccard index values showed a U-shape indicating higher stabilities in Quartiles 1 and 4 compared to Quartiles 2 and 3.

\begin{table}[h]
\caption{Quartile-based BLEU results}
\centering
\begin{tabular}{lccccc}
\toprule
Sample size    & Average & Quartile 1 & Quartile 2 & Quartile 3 & Quartile 4 \\
\midrule
100  & 0.432   & 0.478      & 0.269      & 0.360      & 0.619      \\
200  & 0.464   & 0.521      & 0.313      & 0.380      & 0.643      \\
300  & 0.510   & 0.574      & 0.348      & 0.446      & 0.674      \\
400  & 0.545   & 0.605      & 0.366      & 0.488      & 0.722      \\
500  & 0.557   & 0.594      & 0.387      & 0.509      & 0.739      \\
1000 & 0.644   & 0.657      & 0.476      & 0.624      & 0.818      \\
\bottomrule
\end{tabular}
\label{tab:table4}
\end{table}

\begin{table}[h]
\caption{Quartile-based Jaccard index results}
\centering
\begin{tabular}{lccccc}
\toprule
Sample size    & Average & Quartile 1 & Quartile 2 & Quartile 3 & Quartile 4 \\
\midrule
100  & 0.868 & 0.903 & 0.787 & 0.832 & 0.950 \\
200  & 0.879 & 0.917 & 0.807 & 0.839 & 0.951 \\
300  & 0.897 & 0.930 & 0.825 & 0.864 & 0.970 \\
400  & 0.901 & 0.930 & 0.823 & 0.872 & 0.978 \\
500  & 0.904 & 0.934 & 0.833 & 0.876 & 0.975 \\
1000 & 0.924 & 0.936 & 0.855 & 0.911 & 0.993 \\
\bottomrule
\end{tabular}
\label{tab:table5}
\end{table}

\section{Discussion}
The potential to provide robust explanations is an important desideratum for an explanation tool \cite{lakkaraju20a}. Our empirical study quantifies the stability of SHAP explanations at both the instance (SHAP values) and model-level (variable rankings) and points to a positive relationship between background sample size and the stability of SHAP explanations. More coherent SHAP values and variable rankings were observed when larger background datasets were used. This phenomenon could be partially explained by inference of the central limit theorem: The background dataset converges to the overall distribution at the standard rate of the root of sample size \cite{rosenblatt1956central,jasa}. Therefore, sampling with a larger size could lead to less randomness, the generation of a more representative background dataset, and more stable explanations. Furthermore, our results suggested that the optimal background dataset size depends on a user’s expectation of the ranking exactness. In our pairwise analysis using the BLEU score, we did not observe exact replicate variable order rankings while stable rankings using a fuzzy ranking comparison (Jaccard index) was noted even at small background sizes. Therefore, while SHAP is a trustworthy method for evaluating variable importance, the concrete variable ranking requires careful consideration. Additionally, the U-shape of the comparative BLEU and Jaccard scores indicates that SHAP is more reliable in ranking the most and least important than the moderately important variables.

The results suggest that SHAP users should use a background dataset as large as possible, and could even consider using the whole training dataset as a background dataset. However, larger background datasets lead to more expensive computation and the computing budget is limited for most researchers. To estimate the upper limit of an affordable background sample size, we recommend that SHAP users conduct a pilot experiment using a small background dataset (size of 100, for example) to estimate the computational complexity. Given a complexity $C_{100}$ derived from a background dataset with 100 samples, the complexity $C_m$ using a background dataset with $m$ samples can be approximated by $\frac{m}{100} \times C_{100}$ because of the linear relationship between background sample size and computational complexity \cite{NIPS2017_8a20a862}. Once a decision on the size of the background dataset has been reached, users should make sure that the background dataset is representative for the complete dataset. To this end, some researchers recommend sampling from high-density areas \cite{NIPS2016_5680522b} or using K-means clustering \cite{kmeans}.

There are several limitations to this study. First, it was based on one dataset and DL model. Additional studies using various datasets and DL models can further validate the findings. Second, only 100 simulations were performed in each scenario and future work should include larger simulations. Last, although our study uncovered a stability issue in SHAP explanations and pointed out that large background sizes mitigate the issue, a formal determination of the optimal size remains unexplored.

\section{Conclusion}
Through an empirical study, we have shown that SHAP explanations fluctuate when using a small background dataset and that these fluctuations decrease when the background dataset sampling size increases. This finding holds true for both instance and model-level explanations. Using the BLEU score and Jaccard index for assessing the stability of the model-level variable rankings, we observe a U-shape of the respective measures, which is indicative that SHAP is more reliable in ranking the most and least important than the moderately important variables. Overall, our results suggest that the stability problem caused by the choice of the background dataset size should not be ignored by SHAP users and that users should opt for larger dataset sizes to mitigate fluctuations in SHAP. 

\subsubsection*{URM Statement}
We acknowledge that Han Yuan, Mingxuan Liu, and Chenkui Miao meet the URM criteria.

\bibliography{iclr2023_conference_tinypaper}

\begin{thebibliography}{15}
\providecommand{\natexlab}[1]{#1}
\providecommand{\url}[1]{\texttt{#1}}
\expandafter\ifx\csname urlstyle\endcsname\relax
  \providecommand{\doi}[1]{doi: #1}\else
  \providecommand{\doi}{doi: \begingroup \urlstyle{rm}\Url}\fi

\bibitem[Fletcher \& Islam(2018)Fletcher and Islam]{Fletcher_Islam_2018}
Sam Fletcher and Md~Zahidul Islam.
\newblock Comparing sets of patterns with the jaccard index.
\newblock \emph{Australasian Journal of Information Systems}, 22, Mar. 2018.

\bibitem[Giese et~al.(2021)Giese, Sinn, Wegner, and
  Rappsilber]{giese2021retention}
Sven~H Giese, Ludwig~R Sinn, Fritz Wegner, and Juri Rappsilber.
\newblock Retention time prediction using neural networks increases
  identifications in crosslinking mass spectrometry.
\newblock \emph{Nature Communications}, 12\penalty0 (1):\penalty0 1--11, 2021.

\bibitem[Johnson et~al.(2016)Johnson, Pollard, Shen, et~al.]{johnson2016mimic}
Alistair~EW Johnson, Tom~J Pollard, Lu~Shen, et~al.
\newblock Mimic-iii, a freely accessible critical care database.
\newblock \emph{Scientific data}, 3\penalty0 (1):\penalty0 1--9, 2016.

\bibitem[Kaufmann et~al.(2021)Kaufmann, Lane, Liu, and
  Vecchio]{kaufmann2021efficient}
Kevin Kaufmann, Hobson Lane, Xiao Liu, and Kenneth~S Vecchio.
\newblock Efficient few-shot machine learning for classification of ebsd
  patterns.
\newblock \emph{Scientific reports}, 11\penalty0 (1):\penalty0 1--12, 2021.

\bibitem[Kim et~al.(2016)Kim, Khanna, and Koyejo]{NIPS2016_5680522b}
Been Kim, Rajiv Khanna, and Oluwasanmi~O Koyejo.
\newblock Examples are not enough, learn to criticize! criticism for
  interpretability.
\newblock In D.~Lee, M.~Sugiyama, U.~Luxburg, I.~Guyon, and R.~Garnett (eds.),
  \emph{Advances in Neural Information Processing Systems}, volume~29, 2016.

\bibitem[Lakkaraju et~al.(2020)Lakkaraju, Arsov, and Bastani]{lakkaraju20a}
Himabindu Lakkaraju, Nino Arsov, and Osbert Bastani.
\newblock Robust and stable black box explanations.
\newblock In \emph{International Conference on Machine Learning}, volume 119,
  pp.\  5628--5638, 2020.

\bibitem[Lauritsen et~al.(2020)Lauritsen, Kristensen, Olsen,
  et~al.]{lauritsen2020explainable}
Simon~Meyer Lauritsen, Mads Kristensen, Mathias~Vassard Olsen, et~al.
\newblock Explainable artificial intelligence model to predict acute critical
  illness from electronic health records.
\newblock \emph{Nature Communications}, 11\penalty0 (1):\penalty0 1--11, 2020.

\bibitem[Li \& Ding(2017)Li and Ding]{jasa}
Xinran Li and Peng Ding.
\newblock General forms of finite population central limit theorems with
  applications to causal inference.
\newblock \emph{Journal of the American Statistical Association}, 112\penalty0
  (520):\penalty0 1759--1769, 2017.

\bibitem[Lundberg \& Lee(2017)Lundberg and Lee]{NIPS2017_8a20a862}
Scott~M Lundberg and Su-In Lee.
\newblock A unified approach to interpreting model predictions.
\newblock In \emph{Advances in Neural Information Processing Systems},
  volume~30, 2017.

\bibitem[Lundberg et~al.(2020)Lundberg, Erion, Chen, et~al.]{lundberg2020local}
Scott~M Lundberg, Gabriel Erion, Hugh Chen, et~al.
\newblock From local explanations to global understanding with explainable ai
  for trees.
\newblock \emph{Nature Machine Intelligence}, 2\penalty0 (1):\penalty0
  56--=--67, 2020.

\bibitem[Oba et~al.(2021)Oba, Tezuka, Sanuki, and Wagatsuma]{kmeans}
Yuki Oba, Taro Tezuka, Masaru Sanuki, and Yukiko Wagatsuma.
\newblock Interpretable prediction of diabetes from tabular health screening
  records using an attentional neural network.
\newblock In \emph{International Conference on Data Science and Advanced
  Analytics}, pp.\  1--11, 2021.

\bibitem[Papineni et~al.(2002)Papineni, Roukos, Ward, and
  Zhu]{papineni2002bleu}
Kishore Papineni, Salim Roukos, Todd Ward, and Wei-Jing Zhu.
\newblock Bleu: a method for automatic evaluation of machine translation.
\newblock In \emph{Annual meeting of the Association for Computational
  Linguistics}, pp.\  311--318, 2002.

\bibitem[Rosenblatt(1956)]{rosenblatt1956central}
Murray Rosenblatt.
\newblock A central limit theorem and a strong mixing condition.
\newblock \emph{Proceedings of the National Academy of Sciences}, 42\penalty0
  (1):\penalty0 43, 1956.

\bibitem[van~der Velden et~al.(2020)van~der Velden, Janse, Ragusi,
  et~al.]{van2020volumetric}
Bas~HM van~der Velden, Markus~HA Janse, Max~AA Ragusi, et~al.
\newblock Volumetric breast density estimation on mri using explainable deep
  learning regression.
\newblock \emph{Scientific Reports}, 10\penalty0 (1):\penalty0 1--9, 2020.

\bibitem[Xie et~al.(2022)Xie, Yuan, Ning, et~al.]{XIE2022103980}
Feng Xie, Han Yuan, Yilin Ning, et~al.
\newblock Deep learning for temporal data representation in electronic health
  records: A systematic review of challenges and methodologies.
\newblock \emph{Journal of Biomedical Informatics}, 126:\penalty0 103980, 2022.

\end{thebibliography}
\bibliographystyle{iclr2023_conference_tinypaper}

\end{document}